\documentclass{applemlr}
\usepackage{amsmath}
\usepackage{enumerate} 
\usepackage{algorithm}
\usepackage{algpseudocode}
\usepackage{amsfonts}
\usepackage{amsthm}
\usepackage{newtxtt}
\usepackage{cleveref}
\usepackage{diagbox} 
\usepackage{colortbl}
\usepackage{amssymb}
\usepackage{xspace}
\usepackage{wrapfig}
\usepackage{adjustbox}
\usepackage{tabularx}
\usepackage{booktabs}
\usepackage{mathtools}
\usepackage{tikz}
\usepackage{enumitem}
\usepackage{silence}
\usepackage{dsfont}
\usepackage[table]{xcolor}
\usepackage[dvipsnames]{xcolor}
\usepackage{multirow}
\usepackage{makecell}
\usepackage{xfakebold}
\usepackage{tabularx}

\usepackage{amsmath,amsfonts,bm}

\def\eqref#1{equation~\ref{#1}}

\def\1{\bm{1}}

\DeclareMathAlphabet{\mathsfit}{\encodingdefault}{\sfdefault}{m}{sl}
\SetMathAlphabet{\mathsfit}{bold}{\encodingdefault}{\sfdefault}{bx}{n}

\definecolor{textgray}{HTML}{6E6E73}
\usetikzlibrary{positioning, calc}
\usetikzlibrary{decorations.pathmorphing}

\makeatletter
\patchcmd{\wrong@fontshape}{\@gobbletwo}{}{}{}
\makeatother
\makeatletter
\AtBeginDocument{%
  \urlstyle{sf}
  
}
\makeatother

\numberwithin{equation}{section} 
\tcbuselibrary{minted}
\setminted[python]{
  linenos,
  breaklines,
  fontsize=\footnotesize,
  xleftmargin=2em
}

\definecolor{light}{RGB}{125, 125, 125}
\crefname{tcb@cnt@pbox}{code}{code}
\Crefname{tcb@cnt@pbox}{Code}{Code}
\crefname{assumption}{assumption}{assumption}
\Crefname{assumption}{Assumption}{Assumptions}

\newtcolorbox[auto counter]{pbox}[2][]{
  colback=white,
  title=Code~\thetcbcounter: #2,
  #1,fonttitle=\sffamily,
  fontupper=\sffamily,
  arc=2pt,
  colframe=bgcolor,
  coltitle=fgcolor,
  colbacktitle=bgcolor,
  toptitle=0.25cm,
  bottomtitle=0.125cm
}

\makeatletter
\newcommand\applefootnote[1]{%
  \begingroup
  \renewcommand\thefootnote{}%
  \renewcommand\@makefntext[1]{\noindent##1}%
  \footnote{#1}%
  \addtocounter{footnote}{-1}%
  \endgroup
}
\makeatother

\definecolor{cverbbg}{gray}{0.90}

\title{Pico-Banana-400K: A Large-Scale Dataset for Text-Guided Image Editing}

\author{
\parbox{\textwidth}{
Yusu Qian, Eli Bocek-Rivele, Liangchen Song, Jialing Tong, Yinfei Yang, Jiasen Lu$^\star$, Wenze Hu$^\star$, Zhe Gan$^\star$
}}

\affiliation{Apple}

\contribution{$^\star$Senior authors}

\abstract{
Recent advances in multimodal models have demonstrated remarkable text-guided image editing capabilities, with systems like GPT-4o and Nano-Banana setting new benchmarks. However, the research community's progress remains constrained by the absence of large-scale, high-quality, and openly accessible datasets built from real images. We introduce \textbf{Pico-Banana-400K}, a comprehensive 400K-image dataset for instruction-based image editing.
Our dataset is constructed by leveraging Nano-Banana to generate diverse edit pairs from real photographs in the OpenImages collection. 
What distinguishes Pico-Banana-400K from previous synthetic datasets is our systematic approach to quality and diversity. We employ a fine-grained image editing taxonomy to ensure comprehensive coverage of edit types while maintaining precise content preservation and instruction faithfulness through MLLM-based quality scoring and careful curation.
Beyond single turn editing, Pico-Banana-400K enables research into complex editing scenarios. The dataset includes three specialized subsets: (1) a 72K-example multi-turn collection for studying sequential editing, reasoning, and planning across consecutive modifications; (2) a 56K-example preference subset for alignment research and reward model training; and (3) paired long-short editing instructions for developing instruction rewriting and summarization capabilities.
By providing this large-scale, high-quality, and task-rich resource, Pico-Banana-400K establishes a robust foundation for training and benchmarking the next generation of text-guided image editing models.
}
\metadata[Code]{\url{{https://github.com/apple/pico-banana-400k}}}
\date{\sffamily\today}

\begin{document}

\maketitle

\section{Introduction}
\label{sec:intro}

Recent advances in multimodal large language models (MLLMs) such as GPT-4o~\citep{hurst2024gpt} and Gemini-2.5-Flash-Image (Nano-Banana)~\citep{comanici2025gemini}, along with diffusion-based visual editing models \citep{wu2025qwenimagetechnicalreport,seedream2025seedream40nextgenerationmultimodal,labs2025flux1kontextflowmatching,mou2025instructxunifiedvisualediting}, have demonstrated remarkable capabilities in instruction-guided image editing. These models can transform images based on natural language commands, from simple color adjustments to complex compositional changes.

Despite these advances, open research remains limited by the lack of large-scale, high-quality, and fully shareable editing datasets. 
Existing datasets~\citep{ye2025echo4oharnessingpowergpt4o,hui2024hqedithighqualitydatasetinstructionbased} often rely on synthetic generations from proprietary models or limited human-curated subsets. Furthermore, these datasets frequently exhibit domain shifts, unbalanced edit type distributions, and inconsistent quality control, hindering the development of robust editing models.

To address these challenges, we introduce Pico-Banana-400K, a comprehensive dataset of approximately 400K text-guided image edits built from real photographs in the OpenImages dataset~\citep{krasin2017openimages}. Our dataset represents a systematic effort to create high-quality training data for instruction-based image editing that is both diverse and fully shareable under clear licensing terms.

\begin{figure}[t!]
    \centering
    \includegraphics[width=\textwidth]{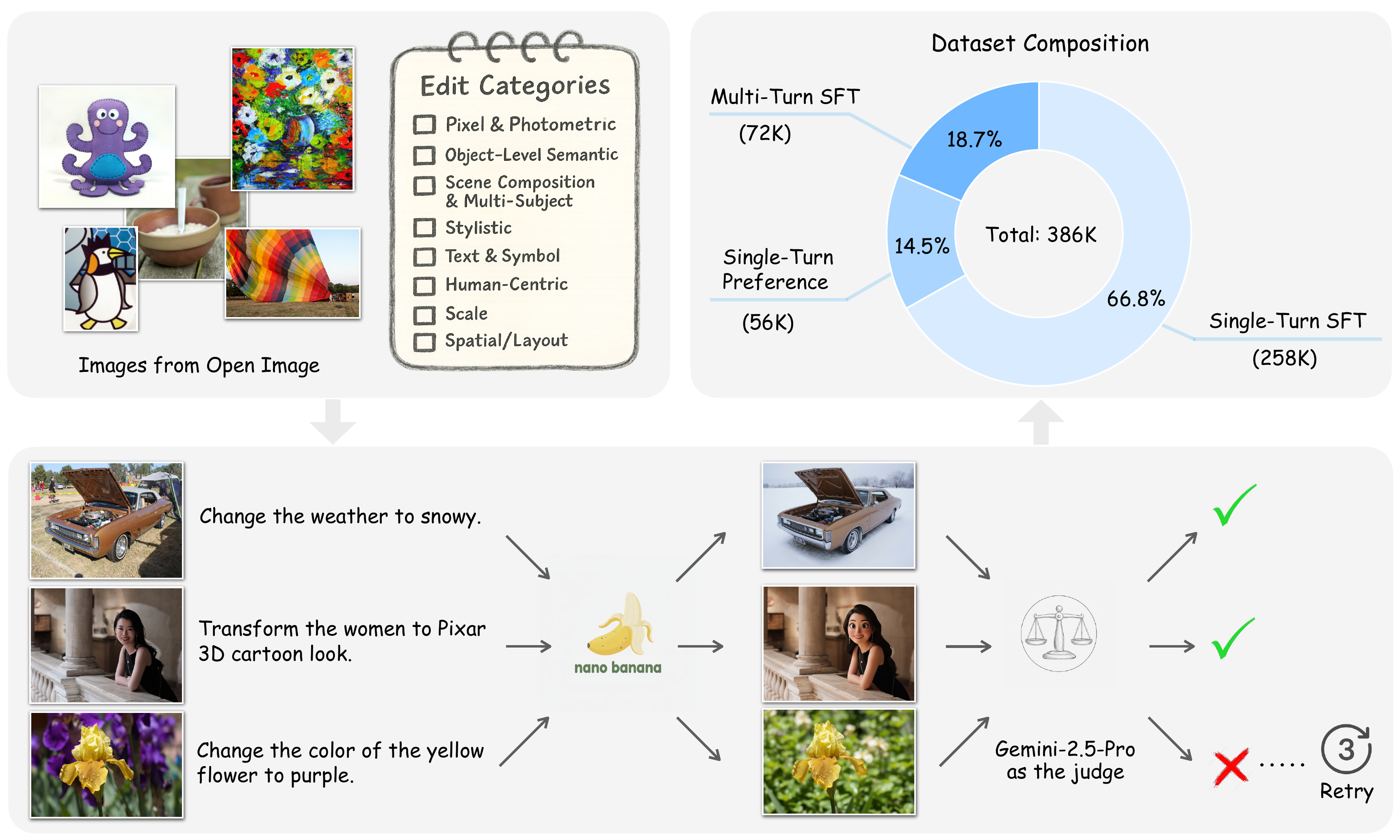}
    \caption{Pico-Banana-400K dataset overview. The pipeline (bottom) shows how diverse OpenImages inputs are edited using Nano-Banana and quality-filtered by Gemini-2.5-Pro, with failed attempts automatically retried. The dataset comprises 386K examples across single-turn SFT (66.8\%), preference pairs (14.5\%), and multi-turn sequences (18.7\%), organized by our comprehensive edit taxonomy (top left).}
    \label{fig:pipeline}
\end{figure}

Figure~\ref{fig:pipeline} illustrates our systematic approach to dataset construction. We leverage Nano-Banana to generate edits across 35 distinct edit types, employ Gemini-2.5-Pro as an automated judge for quality assurance through multi-dimensional scoring (instruction compliance, editing quality, preservation balance, and technical quality), and create specialized subsets for different research needs. Failed editing attempts are automatically retried and preserved as negative examples, while successful edits form our core training data. Additionally, we generate both detailed training-oriented prompts and concise human-style instructions to support diverse research and deployment scenarios. We provide detailed discussion of our construction methodology in Section~\ref{sec:dataset}.

Our contributions are summarized as follows. 
\begin{enumerate}
    \item \textbf{Large-scale shareable dataset}: We release Pico-Banana-400K,\footnote{The total cost of producing this dataset is approximately 100K USD.} containing $\sim$400K high-quality image editing examples built from real images, systematically organized by a 35-type editing taxonomy, with rigorous quality control through automated scoring and manual verification.
    \item \textbf{Multi-objective training support}: Beyond the 258K single-turn supervised fine-tuning examples, we provide 56K preference pairs (successful vs. failed edits) for alignment methods like DPO~\citep{rafailov2024directpreferenceoptimizationlanguage} and reward modeling~\citep{wu2025editrewardhumanalignedrewardmodel}, enabling research on robustness and preference learning.
    \item \textbf{Complex editing scenarios}: We include 72K multi-turn editing sequences where each session contains 2-5 consecutive edits, facilitating research on iterative refinement, context-aware editing, and editing planning. All examples include both detailed and concise instruction variants to study the impact of prompt granularity.
\end{enumerate}
\section{Dataset Construction}
\label{sec:dataset}

\begin{figure*}[t!]
    \centering
    \includegraphics[ width=\textwidth]{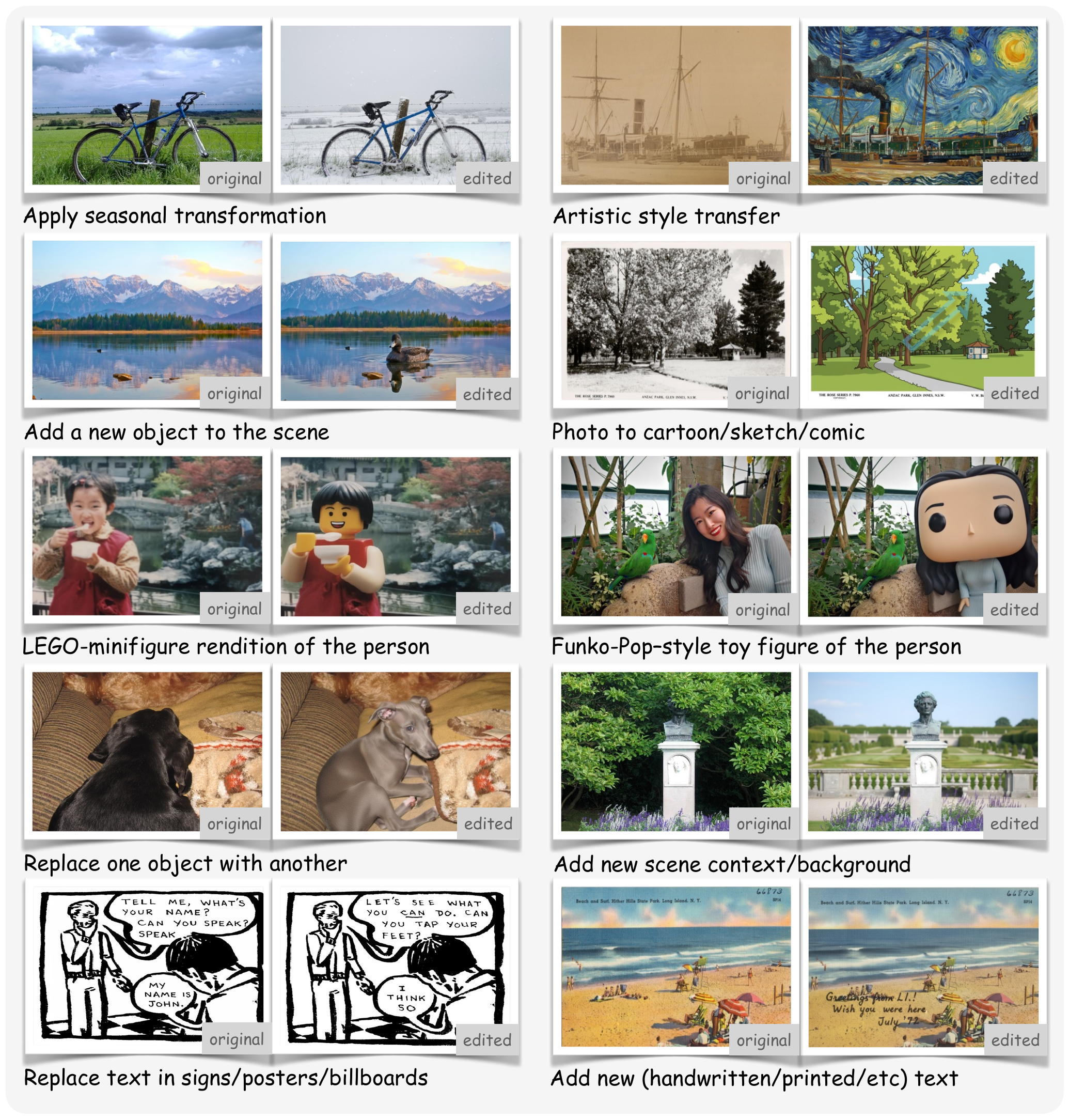}
    \caption{
        \textbf{Example single-turn text-guided image edits from the Pico-Banana-400K dataset.}
        Each pair shows the edited result (right) and its corresponding original image (left).
        The dataset spans diverse edit types, including photometric adjustments, object-level manipulations, stylistic transformations, and scene or lighting modifications.
        These examples illustrate the visual diversity, realism, and high instruction fidelity achieved by the Nano-Banana editing model.
    }
    \label{fig:examples}
\end{figure*}

We construct Pico-Banana-400K through a systematic pipeline designed to ensure both scale and quality. Our approach leverages state-of-the-art models for generation and evaluation while maintaining strict quality control at each stage. We begin by describing our source images and our comprehensive taxonomy of 35 editing operations (Section~\ref{sec:source}). We then detail our dual-instruction generation procedure that creates both detailed training prompts and concise user-style commands (Section~\ref{sec:instruction}). Finally, we present the construction of our single-turn dataset with automated quality assessment (Section~\ref{sec:single_turn}) and our multi-turn editing sequences that enable research on iterative editing scenarios (Section~\ref{sec:multi_turn}).

\begin{table*}[]
\centering
\small
\setlength{\tabcolsep}{6pt}
\begin{tabularx}{\textwidth}{l >{\raggedright\arraybackslash}X r}
\toprule
\textbf{Category} & \textbf{Operation (Edit Type)} & \textbf{Count (Single Turn)} \\
\midrule
\multirow{2}{*}{\textbf{Pixel \& Photometric}}
  & Change overall color tone (warm $\leftrightarrow$ cool) & 14745 \\
  & Add film grain or vintage filter & 15443 \\
\midrule
\multirow{9}{*}{\textbf{Object-Level Semantic}}
  & Add a new object to the scene & 14190 \\
  & Remove an existing object & 15111 \\
  & Replace one object category with another & 14549 \\
  & Change an object's attribute (e.g., color/material) & 13813 \\
  & Relocate an object (change its position/spatial relation) & 6612 \\
  & Change the size/shape/orientation of an object & 10787 \\
\midrule
\multirow{7}{*}{\textbf{Scene Composition \& Multi-Subject}}
  & Add new scene context/background & 14830 \\
  & Apply seasonal transformation (summer $\leftrightarrow$ winter) & 13439 \\
  & Change weather conditions (sunny/rainy/snowy) & 11993 \\
  & Adjust global lighting (e.g., golden hour, fluorescent) & 12433 \\
\midrule
\multirow{4}{*}{\textbf{Stylistic}}
  & Strong artistic style transfer (e.g., Van Gogh/anime/etc.) & 15285 \\
  & Photo $\rightarrow$ cartoon/sketch/comic & 12736 \\
  & Modern $\leftrightarrow$ historical style/look & 14856 \\
\midrule
\multirow{5}{*}{\textbf{Text \& Symbol}}
  & Replace text in signs/posters/billboards & 3495 \\
  & Add new (handwritten/printed/etc.) text & 3867 \\
  & Change font style or color of visible text (if present) & 1432 \\
  & Translate written text into other languages & 1896 \\
\midrule
\multirow{21}{*}{\textbf{Human-Centric}}
  & Add/Remove/Replace accessories (glasses, hats, jewelry, masks) & 1597 \\
  & Clothing edit (change color/outfit) & 1801 \\
  & Pose tweak (minor plausible change) & 1833 \\
  & Modify expressions (smile, frown, neutral) & 1526 \\
  & Change age/gender & 1685 \\
  & Convert person to 2D anime/manga style (identity-preserving) & 1040 \\
  & Convert person to Pixar/Disney-like 3D cartoon look & 1036 \\
  & Convert person to Western comic cel-shaded style & 982 \\
  & Line-art ink sketch of the person & 1482 \\
  & Sticker-ify the person (bold outline, white border) & 1422 \\
  & Caricature with mild feature exaggeration (keep identity) & 832 \\
  & Funko-Pop--style toy figure of the person & 1859 \\
  & LEGO-minifigure rendition of the person & 1568 \\
  & ``Simpsonize'' the person (yellow-skin cartoon style) & 1439 \\
\midrule
\multirow{1}{*}{\textbf{Scale}}
  & Zoom in & 13729 \\
\midrule
\multirow{2}{*}{\textbf{Spatial/Layout}}
  & Outpainting (extend canvas beyond boundaries) & 12403 \\
\bottomrule
\end{tabularx}
\caption{\textbf{Image editing taxonomy.} Each operation is grouped under its category. \textbf{Count} denotes the number of successful samples in the single-turn subset that passed the Gemini-2.5-Pro judge (instruction compliance and visual quality) within at most three retries. If all three attempts fail for an (image, instruction) pair, the case is deemed a failure and discarded from the released set. If one or two attempts before arriving at a successful edit, then the negative edits are also saved to form the preference data.}
\label{tab:edit-taxonomy}
\end{table*}

\subsection{Overview and Edit Taxonomy}
\label{sec:source}

Our dataset is built upon images sampled from OpenImages~\citep{krasin2017openimages}, selected to ensure coverage of humans, objects, and textual scenes. We organize text-guided edits into a comprehensive taxonomy that covers common real-world editing intents while separating local semantic changes from global stylistic or compositional transformations.

Table~\ref{tab:edit-taxonomy} presents our complete taxonomy of 35 edit types across 8 major categories: Pixel \& Photometric, Object-Level Semantic, Scene Composition, Stylistic, Text \& Symbol, Human-Centric, Scale, and Spatial/Layout. Each image-instruction pair is assigned a single primary edit type. For human-centric and text-related operations, we apply category-specific filtering to ensure edits are only attempted on appropriate images.

\textbf{Quality-driven scope decisions.}
During initial construction, we systematically evaluated Nano-Banana's performance across all candidate edit types. We excluded operations that could not be rendered consistently at high quality:
\begin{itemize}[leftmargin=*]
    \item \emph{Adjust brightness/contrast/saturation} and \emph{Sharpen or blur the image}: edits frequently resulted in negligible or unstable visual change relative to the source, reducing supervision signal.
    \item Edits that change the viewer's aspect of a specific object (strong perspective/pose rewrites): prone to structural artifacts.
    \item Two-image composition (merging objects from two different inputs): empirical results were not sufficiently reliable for inclusion as training pairs. 
\end{itemize}

\subsection{Instruction Generation}\label{sec:instruction}

\begin{wrapfigure}{r}{0.5\textwidth}
  \centering
  \includegraphics[width=0.5\textwidth]{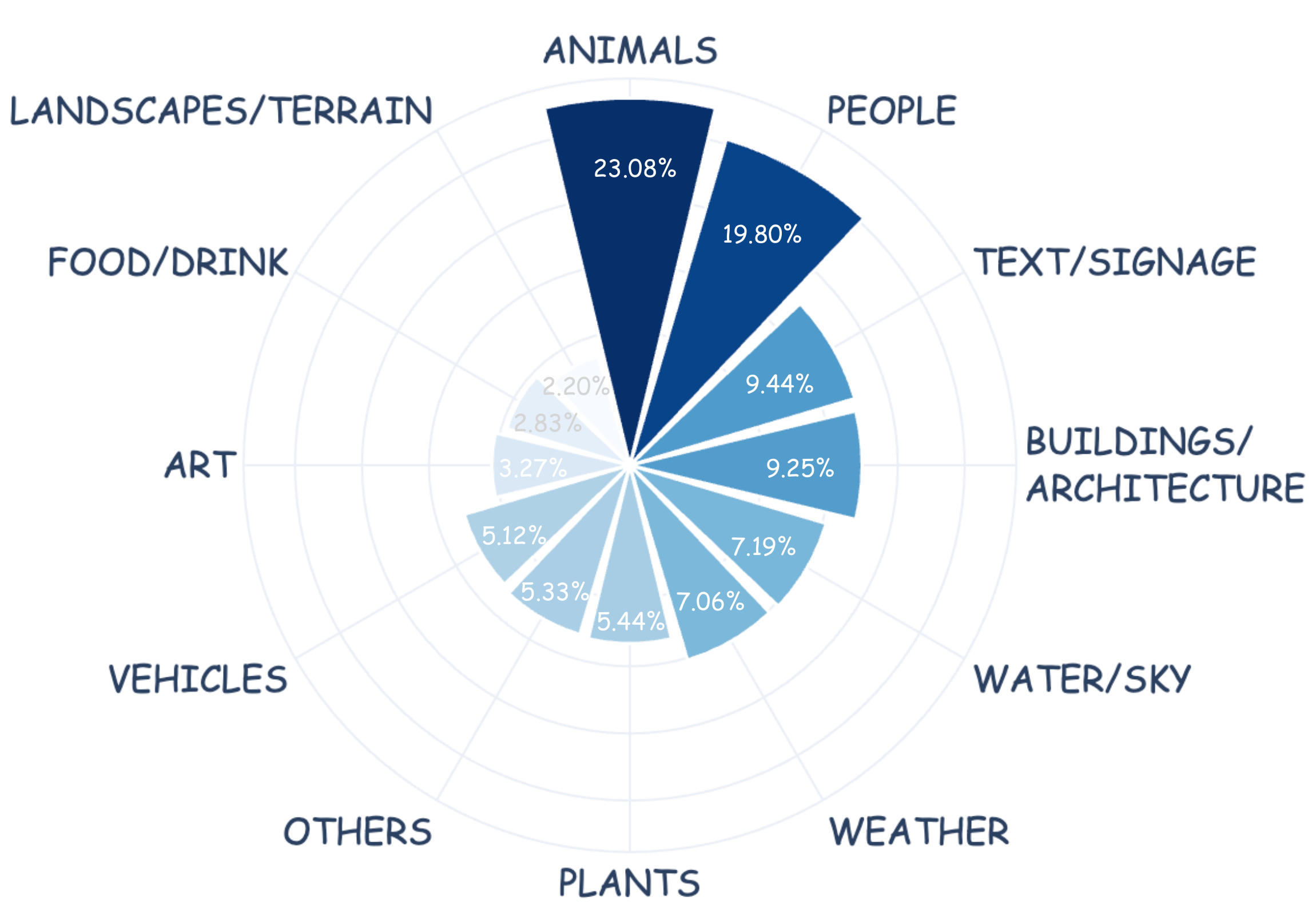}
  \caption{Distribution of image editing instruction content.}
  \label{fig:prompt-content-rose}
\end{wrapfigure}

A key innovation of our dataset is providing dual instruction formats to support diverse research needs. We generate both detailed, training-oriented prompts and concise, human-style commands for each edit.

\textbf{Type I: Long, detailed instructions.}
For each image, we first generate a \emph{long, detailed} editing instruction using Gemini-2.5-Flash with the following system prompt:
\textit{You are an expert photo editor prompt writer.
Given an image, write ONE concise, natural language instruction that a user might give to an image-editing model.
The instruction MUST be aware of visible content (objects, colors, positions) and be closely related to the image content.
Return a JSON object with a ``prompts'' array of photorealistic prompts.}
This version emphasizes unambiguous supervision and is ideal for training setups that benefit from richly specified guidance.

\begin{table}[t]
\centering
\small
\setlength{\tabcolsep}{5pt}
\renewcommand{\arraystretch}{1.25}
\begin{tabular}{p{0.60\linewidth} p{0.36\linewidth}}
\toprule
\textbf{Gemini-generated instruction (long)} & \textbf{Qwen-summarized instruction (short)} \\

\midrule
Reshape the bulky vintage computer monitor on the desk into a slightly more streamlined, less deep CRT model while maintaining its overall screen size and aspect ratio, ensuring the updated form factor casts realistic shadows, reflects ambient light consistently with the scene, and integrates seamlessly with the desk and surrounding environment. &
Reshape the bulky monitor to a sleeker CRT style, keeping the same size and integrating realistically with the desk. \\
\midrule
Replace the current plain sky with a dramatic, modern urban skyline at dusk, featuring sleek glass and steel skyscrapers illuminated with warm interior lights, ensuring the new background's perspective and soft, diffused lighting seamlessly integrate with the existing architectural structure and its upward angle. &
Change the plain sky to a modern urban skyline at dusk with sleek skyscrapers and warm lights, matching the perspective and lighting. \\
\bottomrule
\end{tabular}
\caption{Examples of Gemini written vs.\ Qwen summarized editing instructions.} 
\label{tab:gemini-qwen-top3}
\end{table}

\textbf{Type II: Concise, user-style instructions.}
To study the gap between \emph{model-generated} and \emph{human-like} edit instructions, we launched a focused annotation job to collect \emph{human instructions} for a subset of images. 
We then provide these human-written examples as in-context demonstrations within the system prompt for Qwen2.5-7B-Instruct, which \emph{rewrites} the instructions into a concise, user-style form.
This yields an alternative instruction for the same image/edit intent that better reflects how end-users typically phrase requests. Examples are shown in Table~\ref{tab:gemini-qwen-top3}.

\textbf{Two complementary instruction views.}
Each example in the dataset may therefore contain \emph{two parallel instruction variants}:
(1) a long, detailed instruction from Gemini-2.5-Flash (optimized for data generation and training), and 
(2) a short instruction produced by Qwen using human annotations as examples.
Dataset users can freely choose the variant that best fits their needs (e.g., rich supervision vs.\ natural user prompts).

\textbf{Prompt-derived content distribution.}
To understand which visual domains our editing instructions most frequently target, we categorize each edit instruction into broad image content buckets (e.g., \textsc{People}, \textsc{Animals}, \textsc{Buildings/Architecture}). The categories are inferred via keyword/phrase matching and allow multi-label assignment; for visualization, we aggregate counts per category and render Figure~\ref{fig:prompt-content-rose} which summarizes the content coverage of our prompts.

\begin{figure*}[t!]
  \centering
  \includegraphics[width=\textwidth]{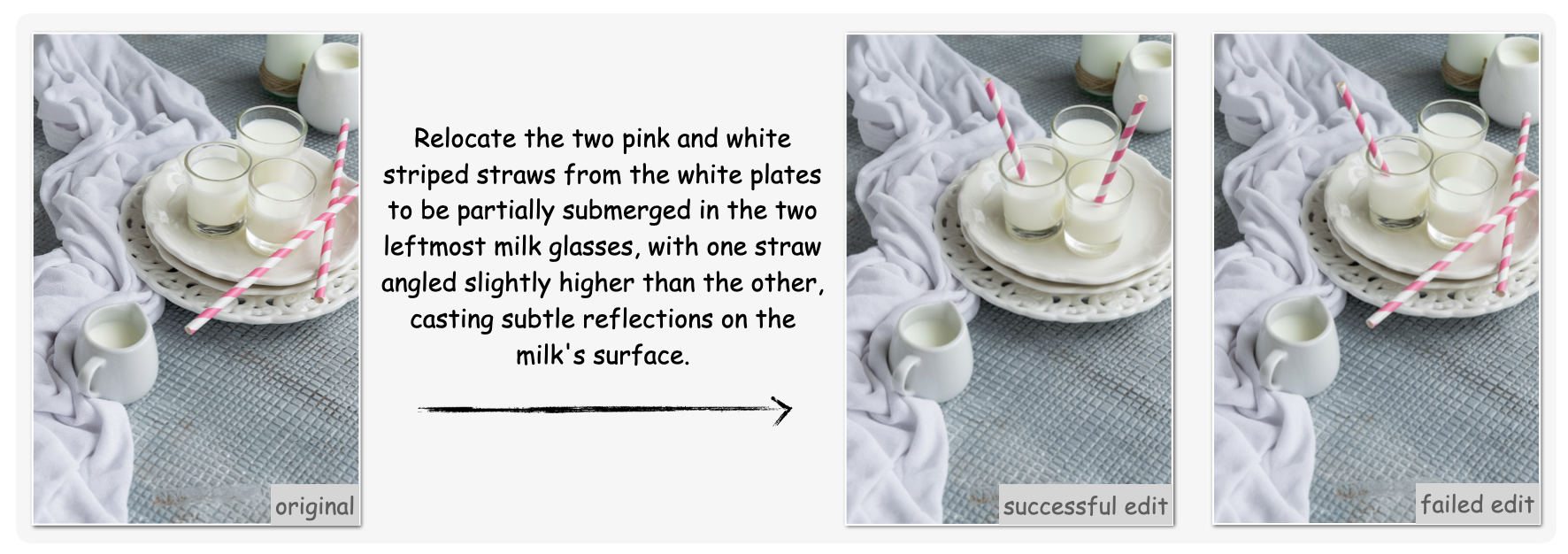}
  \caption{\textbf{Preference triplet example.}
  From left to right: the \emph{original} image, the natural-language \emph{instruction} (center panel) requesting relocation of the pink--white straws into the leftmost glasses, and two model outputs: a \emph{successful edit} that satisfies the instruction and preserves scene context, and a \emph{failed edit} that violates the instruction (incorrect placement/geometry). 
  Such (success, failure) pairs are retained as preference data for alignment studies.}
  \label{fig:pref-triplet}
\end{figure*}

\subsection{Single-Turn Image Editing}
\label{sec:single_turn}
Each edit instruction is executed by Nano-Banana. After generating an edit, Gemini-2.5-Pro serves as an automatic judge that evaluates the edit quality and determines whether it should be retained in the dataset.
The judging process follows a structured system prompt designed to emulate professional human evaluation. The judge evaluates edits using four criteria: Instruction Compliance (40\%), which measures how well the edit fulfills the prompt; Seamlessness (25\%), which checks for natural and artifact-free integration; Preservation Balance (20\%), which ensures unchanged regions remain consistent; and Technical Quality (15\%), which assesses sharpness, color accuracy, and exposure fidelity. We provide the prompt in Appendix~\ref{app:judge_system}.
The resulting score is aggregated into a single quality metric. Images with scores above a strict threshold (empirically set to approximately 0.7) are labeled as successful edits, while those below are categorized as failures. 
\begin{itemize}[leftmargin=*]
    \item \textbf{Successful edits ($\sim$258K)} constitute the main dataset, with examples shown in Figure~\ref{fig:examples};
    \item \textbf{Failure cases ($\sim$56K)} are retained as negative examples paired with successful edits for preference learning. An example triplet is shown in Figure~\ref{fig:pref-triplet}.
\end{itemize}

This self-evaluation process enables Pico-Banana-400K to scale automatically while maintaining high semantic fidelity and visual realism, without requiring human annotators.

\subsection{Multi-Turn Image Editing}\label{sec:multi_turn}

\begin{figure*}[t!]
  \centering
  \includegraphics[width=\textwidth]{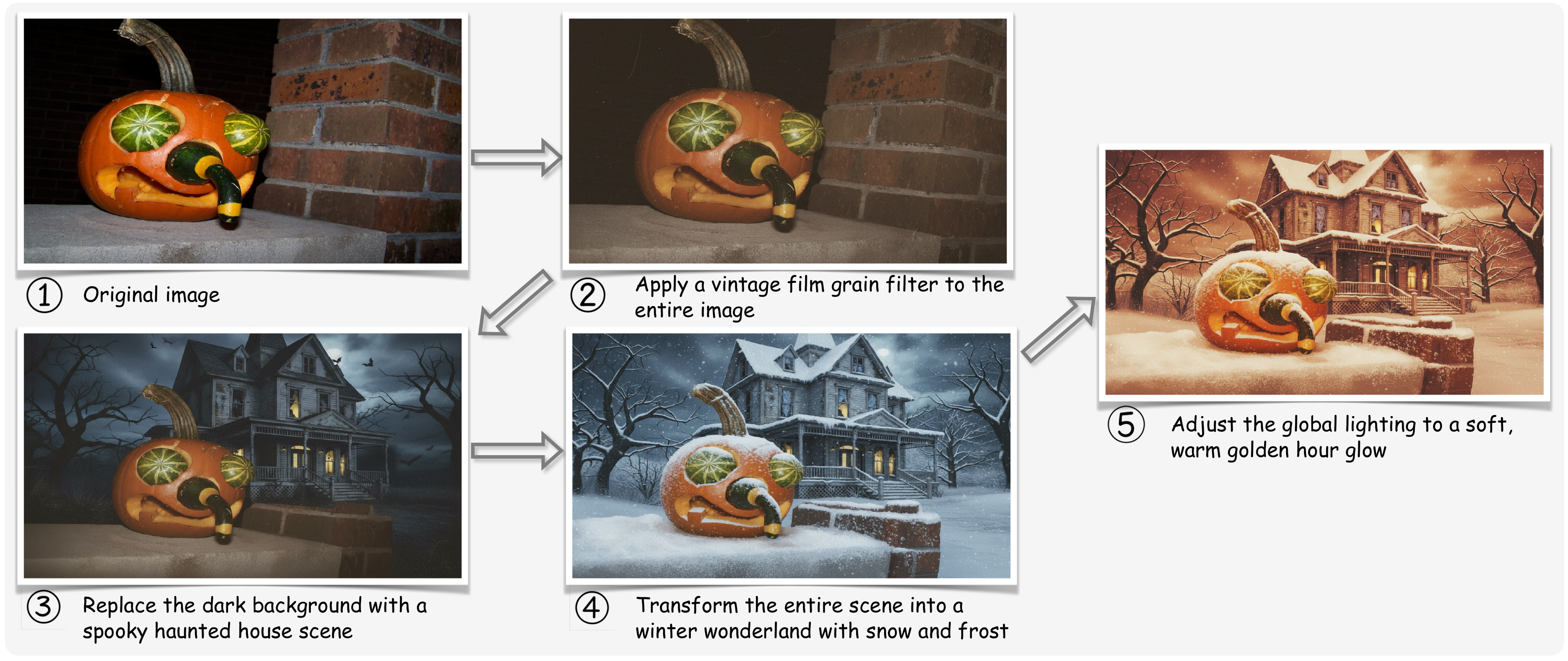}
  \caption{\textbf{Multi-turn image editing example.} Starting from the original pumpkin image, the model first applies a vintage film grain effect, replaces the dark background with a haunted house scene, transforms the entire setting into a snowy winter landscape, and finally adjusts the global lighting to a warm, golden-hour glow, producing the final image on the right.}
  \label{fig:multi-turn example}
\end{figure*}

We build a multi-turn editing subset by expanding a subset of our single-turn editing data. Specifically, we uniformly sample 100K single-turn examples from the dataset introduced earlier. For each sampled example (which already contains its edit type), we create a short editing session by randomly selecting 1–4 additional edit types. This yields sequences of 2–5 total turns per image.

To generate natural, coherent instructions across turns, we prompt Gemini-2.5-Pro to write \textit{single-context} edit instructions conditioned on the image and the history of edit types chosen so far. The model is encouraged to use referential language that links back to prior edits. For instance, if turn~1 is “add a hat to the cat,” turn~2 might say “change the color of \emph{it},” where “it” resolves to the previously added hat. This design emphasizes discourse continuity and dependency between turns rather than independent, disjoint operations.

Execution and evaluation follow the identical procedure used in the single-turn setting: each turn’s instruction is applied to the current working image to produce the next image, and we evaluate the resulting images and instructions with the same criteria and tooling as before. The final dataset therefore provides, for each image, a temporally ordered chain of edits and instructions that exercise both compositionality (multiple edit types) and pragmatic reference (coreference across turns). An example of multi-turn image editing is provided in Figure~\ref{fig:multi-turn example}.
 
\section{Dataset Analysis}

\begin{figure*}[t!]
  \centering
  \includegraphics[width=\textwidth]{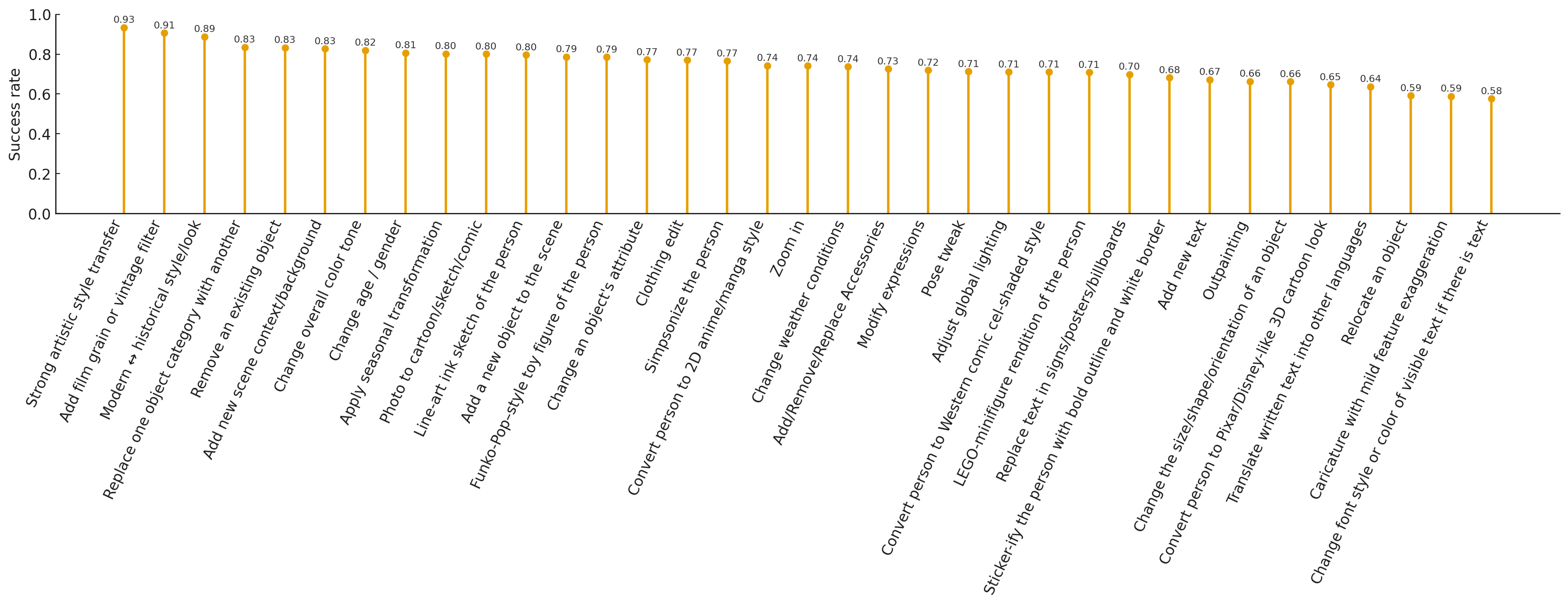}
  \caption{Per–edit type success rates.}
  \label{fig:edit-success-lollipop}
\end{figure*}

We evaluate the success rate of different edit types in our dataset. As shown in Figure~\ref{fig:edit-success-lollipop}, consistent pattern emerges: global appearance and style edits are relatively easy, while edits requiring fine spatial control, layout extrapolation, or symbolic fidelity remain challenging.

\textbf{Easy: global edits and stylization.}
Global edits exhibit the highest reliability. \emph{Strong artistic style transfer} achieves a success rate of $0.9340$, \emph{film grain/vintage} $0.9068$, and \emph{modern$\leftrightarrow$historical restyling} $0.8875$. These operations predominantly reshape global texture, color statistics, and tone, demanding limited spatial reasoning or explicit object coordination.

\textbf{Moderate: object semantics and scene context.}
Semantically targeted, but coarse edits are generally robust. \emph{Remove object} reaches $0.8328$ and \emph{replace category} $0.8348$. Scene-level modifications such as \emph{seasonal change} ($0.8015$) and \emph{photo$\rightarrow$cartoon/sketch} ($0.8006$) perform similarly well. Typical failures stem from imperfect localization under text-only conditioning (e.g., incidental changes to nearby regions) and modest color/texture drift.

\textbf{Hard: precise geometry, layout, and typography.}
Edits requiring fine spatial control or symbolic correctness exhibit the lowest reliability. \emph{Relocate object} is most difficult at $0.5923$, and \emph{change size/shape/orientation} attains $0.6627$, often revealing perspective inconsistencies or topology breaks. \emph{Outpainting} ($0.6634$) struggles with boundary continuity. Text operations are particularly brittle: \emph{change font/style} yields the lowest rate ($0.5759$), while \emph{translate/replace/add text} remain unstable, reflecting challenges in letterform integrity, alignment, and contrast in photorealistic contexts. Among human stylizations, \emph{Pixar/Disney-like 3D} ($0.6463$) and \emph{caricature} ($0.5884$) exhibit identity drift and shading artifacts under large shape exaggerations.

\textbf{Implications.}
Nano-Banana is well suited for \emph{global photometric/stylistic} transformations; in contrast, \emph{fine-grained spatial editing}, \emph{layout extrapolation}, and \emph{typography} remain open problems. Promising directions include stronger spatial conditioning (e.g., region-referential prompting or attention steering), geometry-aware training objectives, explicit text rendering supervision or OCR-informed losses, and identity-preserving constraints for human-centric stylization.

\section{Related Work}
\label{sec:related work}
\textbf{Image Editing Datasets.} Text-guided image editing datasets can be roughly divided into two categories. The first collects paired real image edits with grounded instructions. Prominent examples include GIER~\citep{shi2021learningplanninglanguageguidedglobal} (free-form human-written instructions with before/after pairs) and MagicBrush~\citep{zhang2024magicbrushmanuallyannotateddataset} (10K human-annotated triplets spanning single- and multi-turn edits), then scaling through synthetic or mixed pipelines such as HQ-Edit~\citep{hui2024hqedithighqualitydatasetinstructionbased}, UltraEdit~\citep{zhao2024ultraeditinstructionbasedfinegrainedimage}, OmniEdit~\citep{wei2025omnieditbuildingimageediting}, and UniVG~\citep{fu2025univg}, which expand category coverage, masks, and visual diversity. 

Recently, there is a surging trend to synthesize image editing datasets via distilling frontier multimodal models, e.g., Echo-4o-Image~\citep{ye2025echo4oharnessingpowergpt4o} ($\sim$180K synthetic examples spanning complex-edit generation), and GPT-Image-Edit-1.5M~\citep{wang2025gptimageedit15mmillionscalegptgeneratedimage} (1.5M regenerated triplets unifying OmniEdit/HQ-Edit/UltraEdit). Our dataset also falls into this category, but it is distilled from the most recent Nana-Banana model. A side-by-side comparison of representative image editing datasets is provided in Table~\ref{tab:dataset_comparison}.

\begin{table}[t!]
\centering
\small
\setlength{\tabcolsep}{6pt}
\begin{tabular}{lccc}
\toprule
\textbf{Dataset} & \textbf{Scale} & \textbf{Image Source} & \textbf{Turns} \\
\midrule
GIER~\citep{shi2021learningplanninglanguageguidedglobal} & $10^4$-scale & Real & 1 \\
MagicBrush~\citep{zhang2024magicbrushmanuallyannotateddataset} & $10^4$-scale & Real & 1 / multi \\
HQ-Edit~\citep{hui2024hqedithighqualitydatasetinstructionbased} & $10^5$-scale & Synthetic & 1 \\
Echo-4o-Image~\citep{ye2025echo4oharnessingpowergpt4o} & $10^5$-scale & Synthetic & 1 \\
UltraEdit~\citep{zhao2024ultraeditinstructionbasedfinegrainedimage} & $10^6$-scale & Real & 1 \\
OmniEdit~\citep{wei2025omnieditbuildingimageediting} & $10^6$-scale & Real & 1 \\
GPT-Image-Edit-1.5M~\citep{wang2025gptimageedit15mmillionscalegptgeneratedimage} & $10^6$-scale & Real / Synthetic & 1 \\
\textbf{Pico-Banana-400K (ours)} & $10^5$-scale & Real & 1 / multi \\
\bottomrule
\end{tabular}
\caption{Side-by-side comparison of representative image editing datasets.}
\label{tab:dataset_comparison}
\end{table}

\textbf{Image Editing Models.} Image editing models can be categorized into training-free and finetuning-based approaches. Training-free methods, including foundational diffusion-based techniques like SDEdit~\citep{DBLP:journals/corr/abs-2108-01073}, Prompt-to-Prompt~\citep{ hertz2022prompttopromptimageeditingcross}, and DiffEdit~\citep{couairon2022diffeditdiffusionbasedsemanticimage}, leverage noising–denoising trajectories, attention manipulation, or cross-attention control to enable text-guided edits without retraining. Other notable methods in this category include StableFlow~\citep{avrahami2024stableflow}, FlowEdit~\citep{kulikov2024flowedit}, PnP Inversion~\citep{pnp_inversion}, KV-Edit~\citep{zhu2025kvedittrainingfreeimageediting}, DirectPIE~\citep{ju2023directpie}, and MasaCtrl~\citep{cao_2023_masactrl}. While efficient, these approaches often struggle with complex instructions.

In contrast, finetuning-based methods achieve more precise instruction-following through supervised learning. InstructPix2Pix~\citep{brooks2023instructpix2pix, brooks2023instructpix2pixlearningfollowimage} pioneered this by reformulating editing as learning on (instruction, before, after) triplets. Subsequent models have improved locality, generalization, and multimodal alignment. These include MagicBrush~\citep{zhang2024magicbrush}, which introduced a manually annotated dataset for multi-turn editing; Emu Edit~\citep{sheynin2023emu}, which combines recognition and generation tasks; and others like InstructEdit~\citep{wang2023instructeditimprovingautomaticmasks}, OmniEdit-EditNet~\citep{wei2025omnieditbuildingimageediting}, UltraEdit~\citep{zhao2025ultraedit}, MGIE~\citep{fu2024mgie}, ACE~\citep{han2024ace}, ACE++\citep{mao2025ace++}, SmartEdit\citep{huang2024smartedit}, InsightEdit~\citep{xu2024insightedit}, Qwen-Image-Edit \citep{wu2025qwenimagetechnicalreport}, ICEdit~\citep{zhang2025context}, UniVG~\citep{fu2025univg}, and Step1X-Edit~\citep{liu2025step1x-edit}. The success of these models highlights the value of high-quality, instruction-rich corpora for achieving substantial gains across heterogeneous benchmarks.

\textbf{Positioning.}
Pico-Banana-400K complements prior datasets by emphasizing quality-controlled, instruction-faithful edits and fine-grained category coverage rather than sheer scale. 
It uniquely includes a 56K subset of preference triplets pairing successful and failed edits for alignment research, and a diverse human-centric subset spanning both realistic and stylized transformations—from age or gender changes to anime, Pixar-style, caricature, and LEGO renditions. 
With standardized metadata and ethically sourced imagery, Pico-Banana-400K serves as a large-scale training corpus for text-guided image editing, supporting research on instruction faithfulness and content preservation across edit types.

\vspace{-1mm}
\section{Conclusion}
\vspace{-1mm}
We release Pico-Banana-400K, a large-scale, text-guided image editing dataset aimed to advance image editing research.
By combining Gemini-2.5-Flash for editing instruction generation, Nano-Banana for image editing, and Gemini-2.5-Pro for verification, our work provides a scalable framework for producing high-quality image editing datasets.
All images and metadata are publicly released to support open research in text-guided image editing. Future work includes model benchmarking and model training studies using Pico-Banana-400K, examining how the dataset affects controllability and visual fidelity.

\section*{Acknowledgment}
The authors thank Zhen Yang, Lu Jiang, Chen Chen for valuable guidance, suggestions, and feedback.

\applefootnote{ \textcolor{textgray}{\sffamily Apple and the Apple logo are trademarks of Apple Inc., registered in the U.S. and other countries and regions.}}

\bibliographystyle{plainnat}
\bibliography{biblio}

\clearpage
\appendix
\section{System Prompt for Edit Instruction Generation}

To automatically generate edit instructions for each image, we employed Gemini-2.5-Flash with a carefully designed system prompt that guides the model to behave as a professional photo-editing prompt writer. The model is instructed to produce natural editing instructions that reflect plausible user intents.

The following system-level instruction was provided to Gemini 2.5 Flash:

\begin{quote}
\small
\textbf{System Prompt:} \\
\hspace*{1em}You are an expert photo editor prompt writer.\\
\hspace*{1em}Given an image, write ONE concise, natural language instruction that a user might give to an image-editing model.\\
\hspace*{1em}The instruction MUST be grounded in the visible content (objects, colors, positions) and be closely related to the image content.\\[4pt]
\hspace*{1em}Output Format\\[2pt]
\hspace*{1em}Return a JSON object with a "prompts" array of photorealistic prompts.\\[4pt]
\hspace*{1em}Example Output structure\\
\hspace*{1em}\{\\
\hspace*{2em}"prompts": [\\
\hspace*{3em}"<first prompt>",\\
\hspace*{3em}"<second prompt>"\\
\hspace*{2em}]\\
\hspace*{1em}\}
\end{quote}

\vspace{-2mm}
\section{System Prompt used by Gemini-2.5-Pro as a Judge}
\label{app:judge_system}
\vspace{-2mm}
The system prompt we used to control editing quality is provided as follows:
\begin{quote}
\small
\textbf{System Prompt:} \\
You are a professional image quality evaluator specializing in image editing assessment.

Your task is to evaluate edited images by analyzing the following items in sequence: \\
1. \textbf{Edited Image:} The final edited result (primary evaluation target) \\
2. \textbf{Input Image(s):} One or more reference images used for the edit operation (1--N images) \\
3. \textbf{Editing Instruction:} The specific editing prompt or instruction used

\textbf{Multi-Image Evaluation Context:} You will receive the edited result image first, followed by one or more input images (the reference images used for editing), and finally the editing instruction. Use all of these to make your assessment.

\textbf{Evaluation Criteria (Weighted Scoring for Image Editing):}
\begin{itemize}
    \item \textbf{Edit Instruction Compliance (40\% weight):} Does the edited image fulfill the specific instruction? Are the requested changes clearly visible and properly implemented? Does the result match the intended edit?
    \item \textbf{Editing Quality \& Seamlessness (25\% weight):} Are the edits natural and realistic? Are there visible artifacts, inconsistencies, or blending issues? Is lighting and perspective preserved?
    \item \textbf{Preservation vs. Change Balance (20\% weight):} Are appropriate elements from the original preserved? Are unrelated regions unaffected? Is the editing focused and not overly destructive?
    \item \textbf{Technical Quality (15\% weight):} Overall sharpness, color consistency, exposure, and absence of artifacts or distortions.
\end{itemize}

\textbf{Comparative Analysis:} Compare the edited result against the original image to assess:
\begin{itemize}
    \item What changes were successfully made
    \item What elements were properly preserved
    \item Whether the instruction was accurately interpreted
\end{itemize}

\textbf{Scoring:} Provide a final weighted score from 0.0 to 1.0 based on the evaluation criteria above. The pipeline will automatically compare this score against a strictness threshold.

\end{quote}

\end{document}